\documentclass[sigconf,authorversion,nonacm]{acmart}

\usepackage{multicol}
\usepackage{multirow}
\usepackage{algorithm}
\usepackage{algorithmic}

\AtBeginDocument{%
  }

\setcopyright{acmcopyright}
\copyrightyear{2023}
\acmYear{2023}
\acmDOI{XXXXXXX.XXXXXXX}

\acmConference[WWW 2024]{World Wide Web Conference}{May 13-17, 2024}{Singapore}
\acmPrice{15.00}
\acmISBN{978-1-4503-XXXX-X/18/06}

\begin{document}

\title{MoDE: A Mixture-of-Experts Model with Mutual Distillation among the Experts}

\author{Zhitian Xie}
\authornote{Both authors contributed equally to this research.}
\email{xiezhitian.xzt@antgroup.com}
\affiliation{%
  \institution{Ant Group}
  \streetaddress{569 Xixi Road}
  \city{Hangzhou}
  \country{China}
  \postcode{330100}
}

\author{Yinger Zhang}
\authornotemark[1]
\authornote{All work performed during internship at AntGroup.}
\email{zhangyinger@zju.edu.cn}
\affiliation{%
  \institution{Zhejiang University}
  \city{Hangzhou}
  \country{China}
  \postcode{330100}
}

\author{Chenyi Zhuang}
\authornote{This author is the corresponding author.}
\email{chenyi.zcy@antgroup.com}
\affiliation{%
  \institution{Ant Group}
  \streetaddress{569 Xixi Road}
  \city{Hangzhou}
  \country{China}
  \postcode{330100}
}

\author{Qitao Shi}
\email{qitao.sqt@antgroup.com}
\affiliation{%
  \institution{Ant Group}
  \streetaddress{569 Xixi Road}
  \city{Hangzhou}
  \country{China}
  \postcode{330100}
}

\author{Zhining Liu}
\email{eason.lzn@antgroup.com}
\affiliation{%
  \institution{Ant Group}
  \streetaddress{569 Xixi Road}
  \city{Hangzhou}
  \country{China}
  \postcode{330100}
}

\author{Jinjie Gu}
\email{jinjie.gujj@antgroup.com}
\affiliation{%
  \institution{Ant Group}
  \streetaddress{569 Xixi Road}
  \city{Hangzhou}
  \country{China}
  \postcode{330100}
}

\author{Guannan Zhang}
\email{zgn138592@antgroup.com}
\affiliation{%
  \institution{Ant Group}
  \streetaddress{569 Xixi Road}
  \city{Hangzhou}
  \country{China}
  \postcode{330100}
}

\begin{abstract}
The application of mixture-of-experts (MoE) is gaining popularity due to its ability to improve model's performance.  In an MoE structure, the gate layer plays a significant role in distinguishing and routing input features to different experts. This enables each expert to specialize in processing their corresponding sub-tasks. However, the gate's routing mechanism also gives rise to narrow vision: the individual MoE's expert fails to use more samples in learning the allocated sub-task, which in turn limits the MoE to further improve its generalization ability. To effectively address this, we propose a method called Mixture-of-Distilled-Expert (MoDE), which applies moderate mutual distillation among experts to enable each expert to pick up more features learned by other experts and gain more accurate perceptions on their original allocated sub-tasks. We conduct plenty experiments including tabular, NLP and CV datasets, which shows MoDE's effectiveness, universality and robustness. Furthermore, we develop a parallel study through innovatively constructing "expert probing", to experimentally prove why MoDE works: moderate distilling knowledge can improve each individual expert's test performances on their assigned tasks, leading to MoE's overall performance improvement. 
\end{abstract}

\maketitle
\section{Introduction}

Datasets can be naturally divided into different subsets (such as those from different subdomains or with distinct sub-tasks) and attempting to learn these datasets with a single model may meet difficulties in fitting and generalization \cite{jacobs1991adaptive,eigen2013learning,shazeer2017outrageously}. To address this, the Mixture of Experts (MoE) system has been proposed, which consists of several different experts and a gating network as the router. MoE have been applied in various domains, including multi-task learning \cite{ma2018modeling}, NLP \cite{shazeer2017outrageously,lepikhin2020gshard}, and CV \cite{sparselygatedcnn,pham2021cnn,dosovitskiy2020image, riquelme2021scaling} and proven as a promising architecture. Many studies \cite{jacobs1991adaptive,eigen2013learning,shazeer2017outrageously} have shown that in the MoE structure, each expert is specialized in processing a certain subset of samples.

\begin{figure}[ht]
  \centering
  \includegraphics[width=0.85 \columnwidth]{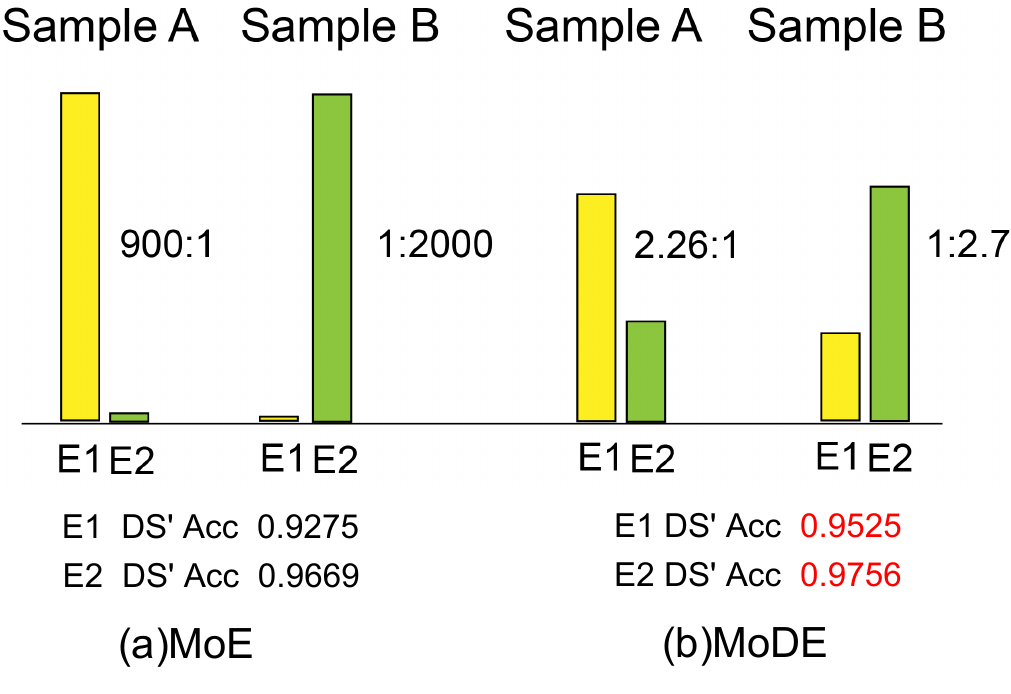}
  \caption{Narrow Vision. A and B represent two subsets of the training samples, while E1/2 represent expert 1/2. The normalized histogram schematically presents the ratio of gradients on each expert in learning the same subset. The narrow vision in MoE is indeed significant and MoDE alleviates it through distillation. As a result, it improves the accuracy of individual experts on their dominating sample-based task domain (referred as \textit{DS} in the following section).} \label{fig1}
\end{figure}

The experts' specialization comes from the fact that they merely learn the limited sample features assigned by the gate during the training process. Figure \ref{fig1}(a) shows that during the training process of an MoE architecture (with totally two experts), different subsets of the training samples partitioned by the gate contribute their learning gradients in significantly different levels: Sample A has a significant gradient to update expert 1's parameters, but has negligible impact on expert 2. Therefore, certain samples contribute to each expert's specialization. However, is the mechanism of allocating limited samples to each expert the best way to construct MoE? Assigning limited samples to each expert in an MoE can lead to a "narrow vision" issue, where experts are not exposed to enough diverse data to develop a comprehensive understanding of their respective sub-tasks, potentially impairing the model's generalization performance.

Can we appropriately address narrow vision in MoE, while still maintaining each expert's specialization to improve the MoE's overall generalization ability? To answer this question, we propose a training methodology called Mixture-of-Distilled-Experts (MoDE), which involves a moderate level of mutual distillation among experts. For each expert, the knowledge obtained from other experts, which is derived from their allocated samples, may provide potentially useful features that are not present in the limited samples allocated to that particular expert. Consequently, this exchange of knowledge can enhance the expert's perception of its sub-task and the MoE's overall performance. Figure \ref{fig1} illustrates the mitigation of narrow vision in MoDE, as evidenced by the increased accuracy of the expert's allocated sub-task (referred as \textbf{\textit{DS}} in the following section). Moreover, plenty of experiments and analysis have been conducted to illustrate the effectiveness and mechanism of MoDE. Innovatively, we conduct a novel parallel study utilizing "expert probing" as an evaluation method to approximate the performance of individual experts within the MoE. We observe that excessive knowledge distillation pushes the experts to present overly similar opinions, undermining the experts' specialization, which deviates from the original motivation of introducing the MoE structure: encourage each expert to be specialized and do what they are good at, resulting in the failure to improve MoE's generalization ability. However, by employing an appropriate distillation strength, each individual expert does not only maintain its specialization but also achieves improved test performance in its allocated sub-task, consequently enhancing the MoE model's generalization ability. With plenty of experiments on the datasets of tabular, NLP and CV, MoDE proves its effectiveness, universality and robustness in solving narrow vision and provide us a valuable exploring space to increase MoE’s generalization ability.

\section{Related Work}
In our work, we apply mutual distillation to the MoE to improve its generalization ability. The most recent related works are introduced in the following subsections.

\subsection{Mixture-of-Experts} \label{21}
MoE was first introduced by Jacob et al. \cite{jacobs1991adaptive} to combine multiple experts, each trained on a different subset of the data, to form a single powerful model. Eigen et al. \cite{eigen2013learning} extends the MoE to a layer in neural network, which consists of a set of experts (neural networks) and a trainable gate. The gate assigns weights to the experts on a per-example basis, which enables the MoE to output a weighted combination of the experts' outputs. As for the gate routing mechanism, dense gate MoE (DMoE) was firstly introduced, which assigns continuous weight to employ all the experts for each input \cite {eigen2013learning,ma2018modeling,jacobs1991adaptive,jordan1994hierarchical,chen1999improved,yuksel2012twenty}. Recently, the sparse gate MoE (SMoE) have been proposed, to reduce the tremendous computational cost because of the enormous parameters, through activating partial experts or subsets of a network for each input \cite{shazeer2017outrageously,zuo2021taming,lewis2021base,fedus2022switch}. 

Although there are many studies on MoE's gate routing strategy, few works on exploring and utilizing the interactions among individual experts, espeically narrow vision, have been conducted. In this study, we discover that the mutual distillation among the experts can facilitate the communication of the knowledge acquired in their distinctive feature learning process, and this exchange yields positive effects on the MoE's overall performance.

\subsection{Knowledge Distillation} \label{sec22}
Knowledge Distillation (KD) is originally proposed by Hinton et al. \cite {hinton2015distilling} to transfer the knowledge from a high-capacity teacher model to a compact student model, aiming at a more efficient deployment of machine learning models on resource-constrained devices. In \cite {hinton2015distilling}, a small model is trained to directly match the output predictions of the teacher model, while in recent years, many methods have been proposed to excavate more information from the teacher, such as intermediate representations, additional attention information \cite{zagoruyko2016paying}, relations of layers \cite{yim2017gift} and mutual relations of data examples \cite{park2019relational}. These methods have broadened the scope of knowledge distillation beyond just directly learning the teacher model's output. Recent advancements in online KD \cite{guo2020online,zhu2018knowledge,kim2021feature,wu2021peer} have enabled updates of the teacher and student models simultaneously, significantly simplifying the training process. Deep mutual learning \cite{zhang2018deep} allows peer students to learn from each other by computing cross-entropy loss between each pair of students. 
Zhu et al. \cite{allen2020towards} investigate the mechanism of ensemble and knowledge distillation. Their work theoretically proves that "multi-view" data structure leads to ensembles provably improving the test performance through learning most of the features.

\section{Preliminary}
\subsection{Mixture-of-Experts} \label{sec31}

\begin{figure}[]
  \centering
  \includegraphics[width=1.0 \linewidth]{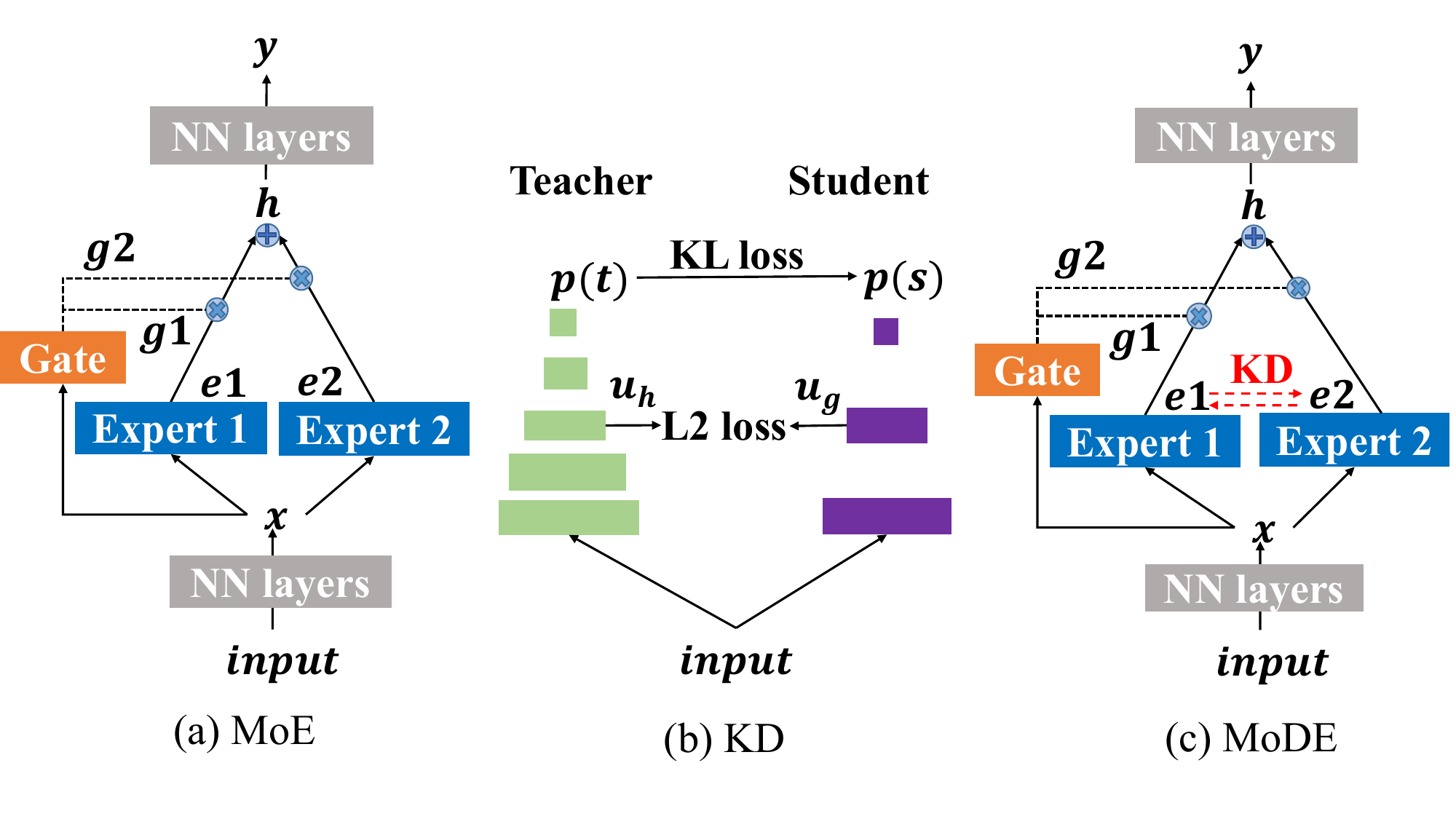}
  \caption{Overview of MoDE.} \label{model}
\end{figure}

Figure \ref{model} (a) illustrates a general MoE layer, which can be inserted into any neural network. The MoE layer accepts the output vector $x$ of the previous layer as input, and outputs a vector $h$ to a successive layer. $y$ is the output of the whole model. The output of MoE layer can be formulated as:

\begin{equation} \label{eq1}
  h= \sum_{i\in \psi }^{} g_{i}(x)e_{i}(x),
\end{equation}
where $e_{i}(x) $ is the output of the $i^{th}$ expert belonging to the expert set $\psi$, $g_{i}(x) $ indicates the weight on the $i^{th}$ expert distributed by the gate. Then the gradient of $h$ with respect to the $i^{th}$ expert's parameters can be written as below:

\begin{equation} \label{eq1_1}
    \frac{\mathrm{\alpha}h}{\mathrm{\alpha}\theta_{i}}=g_{i}(x)\frac{\mathrm{\alpha}e_{i}(x)}{\mathrm{\alpha}\theta_{i}}.
\end{equation}

It indicates that during the training process of each expert, the backpropagation not only depends on the gradient of this expert w.r.t its parameters, but also the gate's weight distribution: a negligible distributed gate weight negligibly updates this expert's parameter using gradient descent.

As introduced in the previous section, MoE can be divided into DMoE and SMoE. DMoE employs all the experts  $N$ for each input, and the gating network $g(x)$ simply comes from a linear transformation $f(.)$ using a softmax layer:

\begin{equation} \label{eq2}
g(x)=Softmax(f(x)), \left | \psi  \right | =N.
\end{equation}
For SMoE, only a part of experts $K$ are selected by the routing strategy. Considering that the routing strategy is not our discussion's topic, so we choose the Topk routing mechanism in \cite{lepikhin2020gshard,fedus2022switch,riquelme2021scaling} as the representative of SMoE, following \cite{riquelme2021scaling}, $g(x)$ can be written as:
\begin{equation} \label{eq3}
g(x)=TopK(Softmax(f(x))+\epsilon ), 
\left | \psi  \right | =K, K\ll N,
\end{equation}
where $\varepsilon \sim N(0,\frac{1}{E^{2} } )$ is the Gaussian noise for exploration of expert routing. When $K\ll N$, most elements of $g(x)$ would be zero so that sparse conditional computation is achieved. 

Various MoE structures applied in our experiments will be discussed later.

\subsection{Knowledge Distillation} 
Figure \ref{model} (b) illustrates two kinds of knowledge distillation methods. In traditional knowledge distillation, the knowledge of distillation comes from the \textit{k} logits output of the model:

\begin{equation} \label{eq4}
p_{i}(z_{i},T)=\frac{exp(z_{i}/T)}{\sum_{i=0}^{k}exp(z_{i}/T)},
\end{equation}

\begin{equation} \label{eq5}
L_{KD}(p(t,T),p(s,T))= {\textstyle \sum_{i=0}^{k}}-p_{i}(t_{i},T)log(p_{i}(s_{i},T)),
\end{equation}

Where T is the temperature coefficient used to control the softening degree of output logit. $p(t,T)$ and $p(s,T)$ represents the soft output logit of the teacher model and student model, respectively. 

Moreover, some studies propose to utilize the features extracted from the teacher model's layer as a guide for the output of the student model's middle layer. We also adopted this approach as the distillation scheme. The L2 loss of the embeddings from the hidden layer is defined as:

\begin{equation} \label{eq6}
L_{KD}(W_{t},W_{s})=\frac{1}{2}\left \| u_{h}(x;W_{t} )-u_{g}(x;W_{s} )   \right \|^{2}.
\end{equation}

$W_{t}$ is the weight of the first h layers of the teacher model, $W_{s}$ is the weight of the first g layers of the student model. $u$ is the output vector of the embedding.

\section{Methodology}

In our work, we propose a methodology called Mixture-of-Distilled-Expert (MoDE), which applies mutual distillation among MoE's experts to encourage each expert to learn more effective features learned by other experts, to gain more accurate perceptions on its learning samples, which in turn increases the overall MoE model's generalization ability. 
We start from an MoDE with 2 experts as a simple example, then we extend it to more experts and sparse gate.

\subsection{Mixture-of-Distilled-Experts} \label{sec32}
Our method establishes a loss function to encourage the knowledge distillation among the experts, denoted as $L_{KD}$. Therefore, the overall loss function $L$ is defined as:
\begin{equation}
L= L_{task}+\alpha L_{KD},  
\end{equation}
where $L_{task}$ represents the loss function of the network itself, which is related to specific tasks, $\alpha$ is the distillation strength and will be discussed in the following section.

When designing the distillation loss $L_{KD}$, we have replaced the conventional teacher-student model with a collaborative learning approach, where each expert functions as a peer student learning from one another. Moreover, we expands the distillation approach from traditionally learning output predictions to intermediate representations. This provides a wider range of potential applications for the proposed MoDE layer, as it can now be easily integrated into any neural network as an independent layer. 

When the expert number in the MoE is $K=2$, the knowledge distillation loss $L_{KD}$ is defined as the squared mean error between the experts' output $e_{1}$ and $e_{2}$. $mean(\cdot)$ operator represents the mean for all dimensions of the vector:

\begin{equation} \label{2 expert equation}
L_{KD} = mean(\left ( e_{1}-e_{2} \right ) ^{2}).
\end{equation}
When the MoE layer is located at the end of the network, $e_1$ and $e_2$ correspond to the output predictions.
While $K>2$, to address the computational complexity, we apply the average of $K$ experts as a single teacher $e_{avg}$ to provide an averaged learning experience and each expert only distills from this single teacher:

\begin{equation}
L_{KD} = \frac{1}{K} \sum_{i=1}^{K} mean(\left ( e_{i} -e_{avg}  \right ) ^{2}) ,
e_{avg}=\frac{1}{K} \sum_{i=1}^{K} e_{i}.
\end{equation}

\subsection{Extension to Sparse-Gated MoE} \label{sec34}
For SMoE, only the activated K experts will be involved in the distillation process, and the method for calculating the distillation loss is the same as in the previous subsection. It is important to note that in DMoE, the outputs $e_{i}$ and $e_{j}$ from expert $i$ and expert $j$ include all samples, whereas in SMoE, a sample only passes through a subset of the experts, and only outputs from the same input can be mutually distilled.

\section{Experiments} \label{sec4}

In this section, we answer the following questions through plenty experiments and discussions: 
\begin{itemize}
\item  \textit{Does MoDE work universally?}
\item  \textit{How and Why MoDE works?}

\item  \textit{Is MoDE robust?}

\end{itemize}

\subsection{Datasets}

\textbf{Tabular Datasets}
7 tabular benchmark data sets of classification task from the OpenML\footnote{https://www.openml.org} are used. Table \ref{tabular datasets} is the basic statistics of the data sets, where $N$, \#$Dim$ and \#$Classes$ are the number of samples, features and classes respectively.

\begin{table}[]\small
\centering
\setlength{\abovecaptionskip}{0.2cm} 
\begin{tabular}{c|ccc}
\hline
Dataset                     & N     & \#Dim & \#Classes \\ \hline
Isolet                      & 7797  & 617   & 26        \\
Mfeat-karhunen              & 2000  & 64    & 10        \\
Mfeat-factors               & 2000  & 216   & 10        \\
Optdigits                   & 5620  & 64    & 10        \\
Satimage                    & 6430  & 36    & 6         \\
First-order-theorem-proving & 6118  & 52    & 6         \\
Artificial-characters       & 10218 & 8     & 10        \\ \hline
\end{tabular}
\caption{Statistics of tabular datasets.}
\label{tabular datasets}
\end{table}

\noindent\textbf{Natural Language Datasets} 
We evaluated our approach on the task of translation, which is widely recognized in the natural language processing. For the low-resource scenario, we used datasets from the IWSLT competitions\footnote{https://iwslt.org/}, specifically the IWSLT14 $English \leftrightarrow German$ ($En \leftrightarrow De$) and IWSLT17 $English \leftrightarrow Arabic$ ($En \leftrightarrow Ar$) translations. The size of the IWSLT datasets are 175k and 241k for $En \leftrightarrow De$ and $En \leftrightarrow Ar$, respectively. For the rich-resource scenario, we used the WMT14 $English \leftrightarrow German$ dataset, which contains approximately 4.5M sentence pairs.

\noindent\textbf{Computer Vision Datasets} 
We apply a variety of datasets: both MNIST \cite{lecun1998gradient}  and Fashion-MNIST \cite{xiao2017fashion} consist of 60,000/10,000 examples of size 28x28 pixels for the training/test set, associated with 10 classes. CIFAR10/100 \cite{krizhevsky2009learning} has a training set with 50,000 images of size 32x32 pixels belonging to 10/100 classes.

\subsection{Models and Settings}
\textbf{Models}
To illustrate various MoE's structures, Table \ref{model name} presents MoE's variations and their corresponding data types. It should be noted that T-DMoE, N-DMoE, and C-DMoE represent three DMoE variations, with the basic DNN structures, Transformer \cite{vaswani2017attention}, and Convolutional Neural Network 
 (CNN) networks, respectively applicable to tabular data, NLP data, and CV data.

In this study, Equation \ref{eq2} is used for DMoE based on previous research \cite{eigen2013learning,ma2018modeling,jacobs1991adaptive,jordan1994hierarchical,chen1999improved}, while the TopK routing mechanism (Equation \ref{eq3}) is applied for SMoE \cite{lepikhin2020gshard,fedus2022switch,riquelme2021scaling}.

T-DMoE and T-SMoE share the same expert structure, with a 2-layer fully-connected (fc) neural network, from input dim to 16 to 10 (equal to class number), where the experts mixture occurs. We take the most popular Transformer network as the backbone architecture for N-DMoE. For the WMT
experiments, the transformer \_vaswani\_wmt\_en\_de\_big setting is used.  Each transformer block consists of a self-attention layer, followed by a feed-forward network and a ReLU non-linearity, namely, FFN. To incorporate MoE models, we replace the FFN layer with MoE models in the 1st encoder and decoder block. Implementation is developed on Fairseq \footnote{https://github.com/facebookresearch/fairseq}.

The design of C-DMoE, modified from \cite{sparselygatedcnn}, utilizes a convolution neural network (CNN) followed by a fc layer to output a 128-dimensional embedding, where expert mixture occurs. Subsequently, the MoE layer's output $h$ is transmitted through a fc layer with a dimension of (128, class). In the case of MNIST and Fashion-MNIST, a 2-layer CNN is utilized, whereas for CIFAR10 and CIFAR100, a ResNet18 \cite{he2016deep} is used.

\noindent\textbf{Settings}
In this work, the number of experts N in all T-DMoE, N-DMoE and C-DMoE is set to 2, and the total number of experts N in T-SMoE is set to 10, while the number of activated experts $K=2$.
The distillation factor $\alpha$ is set to 0.01 or 0.1 in the tablular data sets, 1 in the NLP data sets and 10 in the CV data sets.
Experiments on each dataset are repeated 10 times to calculate the mean and standard variance. All the experiments are conducted on NVIDIA A100 GPUs.

\begin{table}[] \small
\centering
\setlength{\abovecaptionskip}{0.2cm} 
\begin{tabular}{c|ccc}
\hline
\textbf{Name} & \textbf{\begin{tabular}[c]{@{}c@{}}Baseline \\ Architecture\end{tabular}} & \textbf{Gate Type} & \textbf{Data Type} \\ \hline
T-DMoE        & DNN                                                                       & Dense   & Tabular Data       \\
N-DMoE        & Transformer                                                               & Dense   & NLP Data           \\
C-DMoE        & CNN                                                                       & Dense   & CV Data            \\
T-SMoE        & DNN                                                                       & Sparse  & Tabular Data       \\ \hline
\end{tabular}
\caption{MoE's Architectures on Different Datasets.}
\label{model name}
\end{table}

\subsection{Does MoDE Work Universally?} \label{MoDE results}
To verify the effectiveness and universality of MoDE, plenty of experiments has been conducted including the tabular, NLP and CV datasets, where MoE has been widely applied.
 
\subsubsection{Application to Tabular Datasets}
For each tabular data set, we sample a random $60\%$, $20\%$ and $20\%$ of the samples as the training, validation and test set, respectively.

In Table \ref{tabular datasets result}, we present the comparison of various models over 7 multi-category classification tasks. On each dataset, base DMoE and SMoE structures present their advantages over the single model who is identical to the individual expert's architecture. Besides, it can be observed that MoDE with both gate types can give a significantly improved test accuracy than the base models, on all the tabular datasets.

\begin{table*}[]\footnotesize
\centering
\setlength{\abovecaptionskip}{0.2cm}

\begin{tabular}{c|ccc|cc}
\hline
\multirow{2}{*}{Dataset}    & \multirow{2}{*}{Single} & \multicolumn{2}{c|}{Dense-Gated MoE (DMoE)} & \multicolumn{2}{c}{Sparse-Gated MoE (SMoE)} \\ \cline{3-6} 
                            &                         & MoE                  & MoDE                 & MoE                  & MoDE                 \\ \hline
Isolet                      & 0.9228$\pm$0.0064       & 0.9305$\pm$0.0065    & 0.9450$\pm$0.0074    & 0.9426$\pm$0.0035    & 0.9546$\pm$0.0034    \\
Mfeat-karhunen              & 0.9388$\pm$0.0103       & 0.9380$\pm$0.0136    & 0.9525$\pm$0.0079    & 0.9498$\pm$0.0126    & 0.9603$\pm$0.0086    \\
Mfeat-factors               & 0.9617$\pm$0.0080       & 0.9660$\pm$0.0099    & 0.9735$\pm$0.0091    & 0.9662$\pm$0.0090    & 0.9745$\pm$0.0048    \\
Optdigits                   & 0.9658$\pm$0.0081       & 0.9712$\pm$0.0041    & 0.9758$\pm$0.0038    & 0.9760$\pm$0.0035    & 0.9798$\pm$0.0040    \\
Satimage                    & 0.8772$\pm$0.0090       & 0.8872$\pm$0.0097    & 0.8951$\pm$0.0102    & 0.8945$\pm$0.0094    & 0.8970$\pm$0.0085    \\
First-order-theorem-proving & 0.5202$\pm$0.0156       & 0.5269$\pm$0.0209    & 0.5366$\pm$0.0210    & 0.5429$\pm$0.01645   & 0.5611$\pm$0.01499   \\
Artificial-characters       & 0.5994$\pm$0.0076       & 0.6296$\pm$0.0113    & 0.6425$\pm$0.0110    & 0.6619$\pm$0.01372   & 0.6658$\pm$0.01354   \\ \hline
\end{tabular}
\caption{Comparisons on the benchmark tabular datasets. Top-1 Accuracy is reported (Higher is better).}
\label{tabular datasets result}
\end{table*}

\subsubsection{Application to Natural Language Datasets}

All pre-processing steps follow the Fairseq \cite{ott2019fairseq} implementation. We calculate the BLEU scores on these tasks for evaluation \cite{post2018call} and follow the work \cite{vaswani2017attention} for inference. It can be observed in Table \ref{nlp datasets result} that our MoDE model achieves a slight improvement in BLEU score for both low-resource and rich-resource translation tasks, indicating the effectiveness of our proposed method.

\begin{table}[ht]\small
\centering
\setlength{\abovecaptionskip}{0.2cm}

\begin{tabular}{c|ccc}
\hline
Dataset        & Single                                    & MoE                                       & MoDE                           \\ \hline
iwslt 14 de-en & 34.60                                     & 34.88                                     & \textbf{35.14} \\
iwslt 14 en-de & 28.78                                     & 28.66                                     &  \textbf{28.91} \\
iwslt 17-ar-en &  32.92 & 29.81 & \textbf{33.17} \\
iwslt 17-en-ar &  14.42 &  14.46 & \textbf{15.40} \\
wmt14 en-de    & 27.9                                      & 29.10                                     & \textbf{29.48}                        \\ \hline
\end{tabular}
\caption{Comparisons on the benchmark NLP datasets. BLEU is reported (Higher is better).}
\label{nlp datasets result}
\end{table}

\subsubsection{Application to Computer Vision Datasets}
In Table \ref{cv datasets result}, the test accuracy of MoE and MoDE structures (dense gate) have been compared over 4 classic computer vision tasks. On each dataset, although base MoE structures fail to present their advantages over the single model, MoDE can still present a significantly improved test accuracy than both the single model and the corresponding MoE base models.

Through the empirical results, it can be concluded that MoDE proves its effectiveness in enhancing MoE's generalization ability and universality on the tabular, NLP and CV datasets, where MoE architecture has been widely applied.

\begin{table}[]\small
\centering
\begin{tabular}{c|c|ccc}
\hline
Dataset        & Single         & MoE            & MoDE                    \\ \hline
Mnist           & 0.9889 & 0.9886 & \textbf{0.9918} \\
Fashion-M  & 0.8973 & 0.9057 & \textbf{0.9083} \\
Cifar 10          & 0.9462 & 0.9444 & \textbf{0.9519} \\
Cifar 100          & 0.7594 & 0.7545 & \textbf{0.7824} \\ \hline
\end{tabular}
\caption{Comparisons on the benchmark CV datasets. Top-1 Accuracy is reported (Higher is better).}
\label{cv datasets result}
\end{table}

\subsection{How and Why MoDE Works?}
\begin{table*}[ht] \footnotesize
\centering
\setlength{\abovecaptionskip}{0.2cm}

\begin{tabular}{c|ccc}
\hline
Metric                                                     & MoE             & \begin{tabular}[c]{@{}c@{}}MoDE\\ a=0.01\end{tabular} & \begin{tabular}[c]{@{}c@{}}MoDE\\ a=100\end{tabular} \\ \hline
MoE's Overall Performance                                  & 0.9450           & \textbf{0.9625}                                       & 0.9450                                                \\ \hline
MoE's Overall Performance with Gate Masking                & 0.9400            & \textbf{0.9600}                                         & 0.9425                                               \\ \hline
\multirow{2}{*}{Individual Expert's Performance on its DS} & 0.9275 (Exp1)   & \textbf{0.9525 (Exp1)}                                & 0.9600 (Exp1)                                           \\
                                                           & 0.9669 (Exp2)   & \textbf{0.9756 (Exp2)}                                & 0.9590 (Exp2)                                          \\ \hline
Gate's Average Inclination                                 & 0.9496          & \textbf{0.9661}                                       & 0.9108                                               \\ \hline
Gate’s Recognition Accuracy                                & 0.9768(380/389) & \textbf{0.9795(384/392)}                              & 1.000(377/377)                                           \\ \hline
Expert Consistency                                         & 0.0150           & \textbf{0.6125}                                       & 1.000                                                    \\ \hline
\end{tabular}
\caption{How MoDE works? (analysis on Mfeat-karhunen dataset)}
\label{analyze}
\end{table*}

In this subsection, a simplified parallel study has been conducted to illustrate the mechanism and the benefits of MoDE. We use the experiment of DMoE (with two experts) on the Mfeat-karhunen dataset, whose comparison has been demonstrated previously with other datasets.

\textbf{\textit{Definition 1}} When the gate allocates a sample to one expert with more weight (over 0.5 if there are totally two experts) than the other one, then this expert is this sample's \textbf{\textit{dominating expert, DE}}. This sample is in this expert's \textbf{\textit{dominating sample-based task domain, DS}}.

We propose a method called \textbf{\textit{expert probing}}, in approximating each expert's test performance in its \textbf{\textit{{DS}}}.

\textbf{\textit{Methodology: Expert Probing}} For each test sample \textbf{\textit{t}}, there exists a \textbf{\textit{DE}}. By masking this \textbf{\textit{DE}}'s allocated weight to 1, the MoE's prediction output can be compared against \textbf{\textit{t}}'s label. Then the test performance of the \textbf{\textit{DE}} on this \textbf{\textit{t}} can be approximated. Therefore, this individual expert's performance on its \textbf{\textit{DS}} can be evaluated.

Although this mandatory masked output is not exactly the MoE structure's opinion, to some extent, this \textbf{\textit{expert probing}} method can reflect the ability of each individual expert. Moreover, considering that in MoE/MoDE the gate routes a sample to its \textbf{\textit{DE}} with sloping weights (nearly 1) as shown in Table \ref{analyze} and masking the gate brings a negligible effect on the MoE/MoDE's performance, the proposed expert probing method can be regarded as a reliable methodology in estimating each expert's performance.

While analyzing the performance of gate routing, we define a statistic metric \textbf{\textit{Recognition Accuracy}} .

\textbf{\textit{Definition 2}} For one sample, if there exists at least one expert who could correctly predict it (evaluated through \textbf{\textit{expert probing}}), \textbf{\textit{recognition accuracy}} indicates the gate's possibility of correctly picking the expert as the \textbf{\textit{DE}}.

Based on our observation on the test performance of each individual expert, the gate's \textbf{\textit{recognition accuracy}} and their effects on the overall MoE's prediction performance, there are generally two types of errors.
 
\textbf{\textit{Definition 3}} \textbf{\textit{Type 1 Error}} indicates the scenario where the sample cannot be predicted correctly by either individual expert (evaluated through \textbf{\textit{expert probing}}) and the overall MoE architecture cannot thus predict the label correctly; 

\textbf{\textit{Definition 4}} \textbf{\textit{Type 2 Error}} indicates the scenario where the sample can be predicted correctly by at least one individual expert (evaluated through expert probing), however the gate fails to recognize the sample correctly and select the \textbf{\textit{DE}} that cannot predict correctly.

It should be noted that the MoDE model significantly enhances the accuracy over the test dataset with totally 9 more correctly predicted samples. Among them, 6 and 3 samples belong to \textbf{\textit{Type 1 Error}} and \textbf{\textit{Type 2 Error}} observed in the base MoE model, respectively, indicating that MoDE's benefits consists of multiple factors. Table \ref{analyze} has been presented to briefly illustrate each expert's and the gate's performance.

\begin{figure*}[h]
  \centering
  \includegraphics[width=1.0 \linewidth]{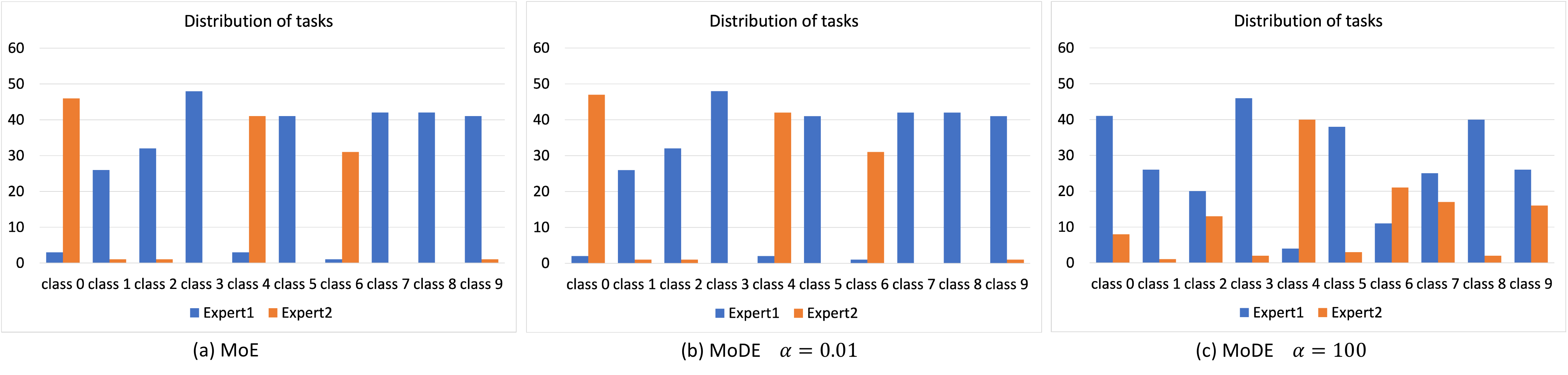}
  \caption{Task Domain Distribution of MoE, MoDE ($\alpha$=0.01), MoDE ($\alpha$=100).} \label{Task Domain Distribution}
\end{figure*}

\subsubsection{The Expert Performs Better in Its \textbf{\textit{DS}}}  During the training process of an MoE model, the cross-entropy loss via gradient descent from random initialization will put a relatively large punishment while the gate allocates a relatively large weight on an expert who however drags the MoE's overall output away from the true label. The gate can thus learn and distinguish the coming samples according to their features and route them to the appropriate experts. Benefiting from this, each expert sees a different subset of training samples and thus obtains its advantage over these samples through training, compared with other experts, which leads to a virtuous coupling effect with the gate's routing mechanism. 
  
   Figure \ref{Task Domain Distribution} shows that each expert of the base MoE obtains its \textbf{\textit{DS}} after training: one expert is specific in the samples generally labeled with overall 7 categories and the other one is specific in the remaining 3 categories. Through \textbf{\textit{Expert Probing}}, the base MoE's individual experts achieve the accuracy of 0.9275 and 0.9669 in their \textbf{\textit{DS}}, respectively. In MoDE with appropriate distillation strength ($\alpha$ = 0.01), each expert does not only maintain its specialization but also significantly raises its accuracy in its \textbf{\textit{DS}} (through \textbf{\textit{Expert Probing}}) to 0.9525 and 0.9756, respectively.

   As mentioned, 6 \textbf{\textit{Type 1 Error}} samples from the base MoE's individual experts are classified correctly by at least one expert in MoDE and thus predicted accurately by the overall MoDE model. Moreover, it should be necessarily noted that each of these 6 samples belongs to its correct \textbf{\textit{DE}}'s original \textbf{\textit{DS}}, according to the labels. In other words, with moderate mutual knowledge distillation, each expert maintains its specialization in its \textbf{\textit{DS}} with increased performance.
  
\subsubsection{The Gate Knows the Experts Better.} 
Table \ref{analyze} shows that in the base MoE model, the gate's \textbf{\textit{recognition accuracy}} reaches 0.9769. As for MoDE ($\alpha$ = 0.01), it raises to 0.9796.

Moreover, it can be found in Table \ref{analyze} that the gate allocates its weights between the experts with enhanced inclination $|g_1-g_2|$ towards each sample, averaging from 0.9496 to 0.9661, which means the gate tends to choose the appropriate expert to handle the current sample more confidently. Due to the enhancement of the gate's \textbf{\textit{recognition accuracy}} in MoDE, 3 \textbf{\textit{Type 2 Error}} samples in the base MoE model can be correctly routed to the correct \textbf{\textit{DE}} by the gate.

\subsubsection{Why Does MoDE Work?}

 As observed in Figure \ref{Task Domain Distribution}, the gate distinguishes the samples' features that implicitly correspond to the labels and then make the routing decision during training. As a result, each expert meets and learns certain features preferred by the gate, towards understanding its \textbf{\textit{DS}}. 
 
 As introduced and proved in Zhu's work \cite{allen2020towards}, each \textbf{\textit{DS}} consists of "multi-view" data structure, where multiple features exist and can be used to classify them correctly and "single-view" data structure, where partial features for the correct labels are missing. During the training process, each expert picks up partial features corresponding to its \textbf{\textit{DS}}, where most multi-view data and merely partial single-view data can be classified correctly and then contribute negligibly, because of the nature of the cross-entropy loss gradient. Due to insufficient amount of left samples, the expert just memorizes instead of learning the remaining single-view training data that the learned features cannot classify correctly. As a result, the individual expert cannot achieve the testing performance as good as training.

 By applying the moderate mutual knowledge distillation to match other experts' output, the training individual expert is prompted to pay attention to other neglected features contained in the remaining single-view samples, although it has already perfectly classified the training data. Therefore, each individual expert presents improved test performance. In light of this, we conduct an experiment and more details are discussed in the next subsection.

\subsubsection{How Do the MoDE's Experts Pay Attention to Other Features?}
We conduct an interpretable model experiment that demonstrates the following: 1. Compared with MoE, MoDE utilizes the input features in a different manner; 2. This distinction can be attributed to the mutual knowledge distillation, where in each individual expert in MoDE pays attention to the features inspired by the other expert(s). In details, we conduct a post-hoc analysis based on our parallel study, to gain a deeper insight on how the trained MoE/MoDE uses the input features to predict the final output.

Considering the complexity of explaining MoE/MoDE, a complicated neural network, we construct an additional agency using the linear regression, which has been widely applied in explaining the neural network, such as Riberio et al.'s work \cite{ribeiro2016should}. Specifically, given the input features, we train a linear regression model to fit MoE/MoDE's logits, based on which each input feature's importance in predicting the logits can be linearly approximated. 

Table \ref{Table 1} displays the 10 most important input features' indexes (from 0 to 63) for each logit (out of 10 categories). It can be observed that towards predicting each logit, MoDE's normalized distribution of the feature weights presents a significant discrepancy from that of MoE, which is quantified as the KL divergence. This indicates that MoDE utilizes these input features in a different manner, which in turn helps it to gain a better test performance that is already proved in the previous subsection.

\begin{table}[h]
\centering
\scriptsize
\caption{MoE/MoDE's top 10 input features for each logit.}
\label{Table 1}
\begin{tabular}{c|ccc}
\hline
\textbf{Dim} & \textbf{Top 10 features in MoE} & \textbf{Top 10 features in MoDE} & \textbf{KL} \\ \hline
0            & 0 10 9 3 1 13 19 11 2 40        & 1 3 14 2 0 9 10 8 27 13       & 0.2038                 \\
1            & 0 4 1 7 6 14 2 8 5 13           & 0 1 4 7 6 14 2 13 5 11        & 0.2033                 \\
2            & 5 0 4 19 24 13 8 7 2 14         & 5 1 4 17 6 7 13 2 39 31       & 0.5120                 \\
3            & 14 0 27 3 7 18 12 10 24 8       & 14 3 7 18 27 26 50 10 12 11   & 0.3194                 \\
4            & 0 1 14 3 27 22 24 8 7 17        & 3 14 1 0 27 7 22 8 12 17      & 0.1819                 \\
5            & 0 1 22 5 4 2 24 9 7 55          & 0 1 22 4 5 2 8 7 50 18 33     & 0.1799                 \\
6            & 5 1 19 24 54 42 8 25 38 22      & 5 0 6 1 3 22 38 33 54 8       & 0.2604                 \\
7            & 4 0 7 20 6 2 12 33 17 8         & 0 7 4 6 2 33 9 14 17 23       & 0.2013                 \\
8            & 0 1 22 26 24 35 18 4 7 45       & 0 22 1 18 2 26 16 35 45 7     & 0.1604                 \\
9            & 0 9 19 24 11 4 13 20 10 54      & 9 4 20 11 2 6 0 19 10 1       & 0.3126                 \\ \hline
\end{tabular}
\end{table}

Furthermore, we conduct a micro-analysis to understand how each individual expert in MoE/MoDE utilizes the input features to express its opinion. Again, we apply the linear regression to approximate the weight of each input in determining each entry of the individual expert's final layer. These approximations are then aggregated across the layer to determine the overall importance of each input feature to the individual expert. We present the rankings of these input features' indexes in Table 8, where higher absolute weights indicate higher prioritization. As discussed, each expert in MoDE maintains its \textit{\textbf{DS}} alongside its corresponding MoE version. Therefore, our focus is to examine the discrepancies in the distribution of input weights between each individual expert's MoE and MoDE version. For instance, there are totally 26 input features (out of 64) whose weight rankings increase in MoDE's expert 1, compared to its MoE version, 13 of which are directly influenced by expert 2, as they have higher rankings in expert 2 compared to expert 1 in the MoE version. The rest 13 input features are raised indirectly, due to the implicit dependencies among the input features. As for MoDE's expert 2, there are totally 35 input features (out of 64) with enhanced weight rankings, 26 of which are directly raised by expert 1. Moreover, the KL divergence between the individual experts' feature importance are presented in Table \ref{Table 3}. In conclusion, being benefited from the mutual knowledge distillation, one expert perceives the other expert's opinion that is implicitly expressed through its input feature weights, which in turn requires this expert to pay more attention to the features previously ignored by itself. As a result, the ignored features that prove their superior importance in helping this expert in learning its \textit{\textbf{DS}} are allocated with more weights and relatively higher rankings. 

\begin{table}[h]
\scriptsize
\centering
\label{Table 2}
\caption{Input's importance ranking to MoE/MoDE's expert.}
\begin{tabular}{cc}
\hline
Expert 1 in MoE                                                                                                                                                                                                                            & Expert 2 in MoE                                                                                                                                                                                                                             \\ \hline
\begin{tabular}[c]{@{}c@{}}0 1 4 5 20 7 18 2 14 6 22 26 9 35 12 \\ 13 24 3 33 11 16 17 37 10 52 55 21 \\ 27 50 42 32 31 30 39 28 45 44  8 40 \\ 19 25 54 57 36 53 43 58 23 46 56 59  \\ 41 38 2 49 15 60 47 51 34 48 62 \\ 61 63\end{tabular}  & \begin{tabular}[c]{@{}c@{}}1 3 0 5 17 10 6 14 42 25 22 4 24 47 \\ 9 7 63 27 40 18 2 36 57 11  61 21 39 \\ 60 38 28 46 41 13 44 26 48  8 54 50 \\ 35 16 62 52 51 23 33 19 32 30 37 53 \\ 43 20 55 49 29 45 15 58 56 59 31 \\ 12 34\end{tabular} \\ \hline
Expert 1 in MoDE                                                                                                                                                                                                                           & Expert 2 in MoDE                                                                                                                                                                                                                            \\ \hline
\begin{tabular}[c]{@{}c@{}}0 1 4 5 14 7 6 2 9 18 20 22 26 13 \\ 11 17 35 3 8 16 12 24 37 33 50 21 \\ 40 27 45 25 10 32 31 29 55 23 19 30 \\ 44 42 39 48 28 52 54 58 56 53 41 \\ 36 49 59 34 15 57 60 38 61 43 47 51 \\ 63 62 46\end{tabular} & \begin{tabular}[c]{@{}c@{}}0 1 14 10 3 6 22 7 9 5 4 17  2 8 21 \\ 28 25 50 33 24 26 32 27 11 42 16 19 \\ 12 13 55 23 18 29 38 20 63 35 30 44 \\ 52 40 60 54 45 49 56 61 36  53 41 39 \\ 48 51 58 62 31 43 59 47 34 46 37 \\ 57 15\end{tabular} \\ \hline
\end{tabular}
\end{table}

\begin{table}[h]
\centering
\footnotesize
\caption{The KL divergence of experts' feature weights.}
\label{Table 3}
\begin{tabular}{cc}
\hline
    \textbf{Expert pair}                       & \textbf{KL divergence} \\ \hline
Expert 1 in MoE vs Expert 1 in MoDE   & 0.0111                 \\
Expert 2 in MoE vs Expert 2 in MoDE   & 0.0623                 \\
Expert 1 in MoE vs Expert 2 in MoE     & 0.1138                 \\
Expert 1 in MoDE vs Expert 2 in MoDE & 0.0474                 \\ \hline
\end{tabular}
\end{table}

 \subsubsection{Will the Experts Degenerate into the Same Network?} 

 There is a concern regarding whether MoDE forces the experts to be similar and deviates the original motivation of introducing MoE: encourage each expert to be specialized and do what they are good at. To explore this, an overly heavy knowledge distillation strength ($\alpha$ = 100) is placed to force both experts to express the exact opinion towards each sample. Figure \ref{Task Domain Distribution} shows that there is a significant shift in these individual experts' \textbf{\textit{DS}} and both experts presents the totally same opinion towards each sample (consistency equals to 1) as shown in Table \ref{analyze}. Although the gate's \textbf{\textit{recognition accuracy}} is raised to 1, MoDE fails to achieve a higher test accuracy, because each individual expert fails to maintain its \textbf{\textit{DS}} and gains better test performance in it. While with a moderate knowledge distillation strength (for example, $\alpha$ = 0.01), although the individual experts' consistency is 0.6125, much higher that of base MoE (0.015), both experts still maintains their corresponding \textbf{\textit{DS}}, which means each individual expert is still specialized. Furthermore, each expert also gains better perception in its \textbf{\textit{DS}} as shown in Table \ref{analyze}. As a result, with improved gate's \textbf{\textit{recognition accuracy}}, MoDE with moderate knowledge distillation strength shows its superior generalization. Further discussions about the knowledge distillation strength are presented in the next subsection.

 \subsection{Ablation Study} \label{sec45}

\subsubsection{More Individual Experts Employed by MoDE}
Table \ref{expert number} selects tabular datasets as the benchmark to analyze the effect of the expert's amount and shows that both MoE and MoDE benefit from the increasing expert's amount with improved test performance. Moreover, the MoDE still maintains a higher accuracy than the base MoE employing the same number of experts, which means the mechanism of mutual knowledge distillation among experts works, regardless of the number of sub-networks employed.

\begin{table*}[ht]\small
\centering
\begin{tabular}{ccc|cc|cc}
\hline
\multirow{2}{*}{Dataset}    & \multicolumn{2}{c|}{2 Experts}           & \multicolumn{2}{c|}{4 Experts}           & \multicolumn{2}{c}{8 Experts}            \\ \cline{2-7} 
                            & MoE            & MoDE             & MoE            & MoDE             & MoE            & MoDE             \\ \hline
Isolet                      & 0.9305(0.0065) & \textbf{0.9450(0.0074)} & 0.9352(0.0068) & \textbf{0.9483(0.0059)} & 0.9410(0.0076) & \textbf{0.9522(0.0046)} \\
Mfeat-karhunen              & 0.9380(0.0136) & \textbf{0.9525(0.0079)} & 0.9432(0.0109) & \textbf{0.9543(0.0109)} & 0.9433(0.0096) & \textbf{0.9602(0.0083)} \\
Mfeat-factors               & 0.9660(0.0099) & \textbf{0.9735(0.0091)} & 0.9620(0.0057) & \textbf{0.9742(0.0040)} & 0.9617(0.0060) & \textbf{0.9755(0.0058)} \\
Optdigits                   & 0.9712(0.0041) & \textbf{0.9758(0.0038)} & 0.9711(0.0025) & \textbf{0.9762(0.0042)} & 0.9737(0.0060) & \textbf{0.9781(0.0043)} \\
Satimage                    & 0.8872(0.0097) & \textbf{0.8951(0.0102)} & 0.8925(0.0117) & \textbf{0.8950(0.0093)} & 0.8990(0.0079) & \textbf{0.9005(0.0088)} \\
1st-order-theorem & 0.5269(0.0209) & \textbf{0.5366(0.0210)} & 0.5315(0.0193) & \textbf{0.5388(0.0158)} & 0.5370(0.0151) & \textbf{0.5450(0.0163)} \\
Artificial-characters       & 0.6296(0.0113) & \textbf{0.6425(0.0110)} & 0.6477(0.0079) & \textbf{0.6568(0.0163)} & 0.6540(0.0167) & \textbf{0.6644(0.0110)} \\ \hline
\end{tabular}
\caption{MoDE's robustness w.r.t more individual experts.}
\label{expert number}
\end{table*}

\subsubsection{Distillation Strength} \label{sec452}

\begin{figure}[h]
  \centering
  \includegraphics[width= 0.85 \linewidth]{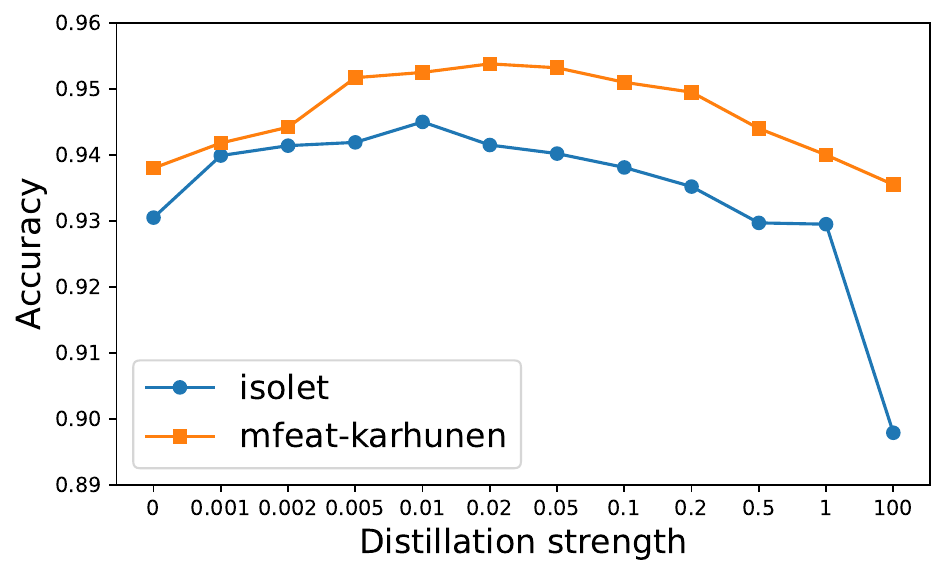}
  \caption{MoDE w.r.t the
distillation strength $\alpha$.} \label{ablation-regulation}
\end{figure}

To investigate the effect of the distillation strength $\alpha$, we run experiments
on the tabular dataset (Isolet and Mfeat-karhunen, considering the distillation strength's scope consistency), which can be found in Figure \ref{ablation-regulation}. $\alpha$ equivalent to zero stands for the test performance of the base MoE model. As $\alpha$ keeps increasing and surpasses a certain point, it tends to push the experts to express overly similar opinions that have been discussed previously and fails to improve the MoE's test accuracy. The relatively wide range that the appropriate distillation strength $\alpha$ lays in maintains MoDE's advantage over the base MoE model and presents its robustness to provide us a valuable exploring space in increasing MoE's generalization ability.

\section{Conclusions}
In this work, we introduce narrow vision, where each individual MoE's expert fails to use more samples in learning the allocated sub-task and thus limits the overall MoE's generalization. To address this, we propose Mixture-of-Distilled-Expert (MoDE), which applies moderate mutual distillation among the experts to encourage them to gain more accurate perceptions on their corresponding distributed tasks. 

Through "expert probing", an innovative evaluation method proposed by us, we find that excessive distillation pushes the experts to presents overly similar opinions, which deviates the original motivation of  MoE's structure and thus fails to improve its generalization ability. However, with moderate distillation strength, each individual expert does not only maintain its specialization in its \textbf{\textit{DS}} but also gains improved test performance along with the gate, where an analysis has been experimentally and analytically conducted. Moreover, this moderate distillation strength lays in a relatively wide range, presenting the robustness. 

With plenty of experiments on the datasets of tabular, NLP and CV, MoDE proves its effectiveness, universality and robustness in solving narrow vision and provide us a valuable exploring space in increasing MoE's generalization. In our future work, we plan to implement our proposed methodology in the most recent industrial scenarios, such as Large Language Model.

\clearpage

\bibliography{aaai24}

\end{document}